\documentstyle[epsf,aclap]{article}       
\include{psfig}

\def\epsfsize#1#2{\ifnum#1>\hsize\hsize\else#1\fi}

\title{\small In {\it Proceedings of the Second Conference on
Empirical Methods in Natural Language Processing, 1997.}
\\ ~ \\
\Large {\bf A Corpus-Based Approach for Building Semantic Lexicons}}

\date{}

\author{Ellen Riloff and Jessica Shepherd \\
Department of Computer Science \\
University of Utah \\
Salt Lake City, UT 84112 \\
{\tt riloff@cs.utah.edu}
}

\begin{document}
\bibliographystyle{fullname}  
\maketitle

\begin{abstract} 
Semantic knowledge can be a great asset to natural language processing
systems, but it is usually hand-coded for each application. Although
some semantic information is available in general-purpose knowledge
bases such as WordNet and Cyc, many applications require
domain-specific lexicons that represent words and categories for a
particular topic.  In this paper, we present a corpus-based method
that can be used to build semantic lexicons for specific
categories. The input to the system is a small set of seed words for a
category and a representative text corpus. The output is a ranked list
of words that are associated with the category. A user then reviews
the top-ranked words and decides which ones should be entered in the
semantic lexicon.  In experiments with five categories, users
typically found about 60 words per category in 10-15 minutes to build a
core semantic lexicon.
\end{abstract}

\section{Introduction}

Semantic information can be helpful in almost all aspects of natural
language understanding, including word sense disambiguation,
selectional restrictions, attachment decisions, and discourse
processing. Semantic knowledge can add a great deal of power and
accuracy to natural language processing systems. But semantic
information is difficult to obtain. In most cases, semantic knowledge
is encoded manually for each application.

There have been a few large-scale efforts to create broad semantic
knowledge bases, such as WordNet~\cite{miller90} and
Cyc~\cite{aimag86}. While these efforts may be useful for some
applications, we believe that they will never fully satisfy the need
for semantic knowledge.  Many domains are characterized by their own
sublanguage containing terms and jargon specific to the
field. Representing all sublanguages in a single knowledge base would
be nearly impossible. Furthermore, domain-specific semantic lexicons
are useful for minimizing ambiguity problems. Within the context of a
restricted domain, many polysemous words have a strong preference for
one word sense, so knowing the most probable word sense in a domain
can strongly constrain the ambiguity.

We have been experimenting with a corpus-based method for building
semantic lexicons semi-automatically. Our system uses a text corpus
and a small set of seed words for a category to identify other words
that also belong to the category. The algorithm uses simple statistics
and a bootstrapping mechanism to generate a ranked list of potential
category words. A human then reviews the top words and selects the
best ones for the dictionary. Our approach is geared toward fast
semantic lexicon construction: given a handful of seed words for a
category and a representative text corpus, one can build a semantic
lexicon for a category in just a few minutes.

In the first section, we describe the statistical bootstrapping
algorithm for identifying candidate category words and ranking
them. Next, we describe experimental results for five
categories. Finally, we discuss our experiences with additional
categories and seed word lists, and summarize our results.

\section{Generating a Semantic Lexicon}
\label{algorithm-section}

Our work is based on the observation that category members are often
surrounded by other category members in text, for example in
conjunctions ({\it lions and tigers and bears}), lists ({\it lions,
tigers, bears...}), appositives ({\it the stallion, a white Arabian}),
and nominal compounds ({\it Arabian stallion}; {\it tuna fish}). Given
a few category members, we wondered whether it would be possible to
collect surrounding contexts and use statistics to identify other
words that also belong to the category. Our approach was motivated by
Yarowsky's word sense disambiguation algorithm~\cite{yarowsky92} and
the notion of statistical salience, although our system uses somewhat
different statistical measures and techniques.

We begin with a small set of seed words for a category.  We
experimented with different numbers of seed words, but were surprised
to find that only 5 seed words per category worked quite well. As an
example, the seed word lists used in our experiments are shown below.

\begin{figure}[hbt]
\fbox{
\begin{minipage}{2.9in}
\begin{tabular}{ll}
\hspace*{-.15in} {\bf Energy:} & \hspace*{-.15in} {\it fuel gas gasoline
oil power} \\ 
\hspace*{-.15in} {\bf Financial:} & \hspace*{-.15in} {\it bank banking currency dollar money} \\ 
\hspace*{-.15in} {\bf Military:} & \hspace*{-.15in} {\it army commander infantry soldier} \\
~ & \hspace*{-.15in} {\it troop} \\ 
\hspace*{-.15in} {\bf Vehicle:} & \hspace*{-.15in} {\it airplane car jeep plane truck} \\ 
\hspace*{-.15in} {\bf Weapon:} & \hspace*{-.15in} {\it bomb dynamite explosives gun rifle}
\end{tabular}
\end{minipage}
}
\caption{Initial Seed Word Lists}
\label{seed-words}
\end{figure}

The input to our system is a text corpus and an initial set of seed
words for each category. Ideally, the text corpus should contain many
references to the category. Our approach is designed for
domain-specific text processing, so the text corpus should be a
representative sample of texts for the domain and the categories
should be semantic classes associated with the domain. Given a text
corpus and an initial seed word list for a category C, the algorithm
for building a semantic lexicon is as follows:

\begin{enumerate}
\item We identify all sentences in the text corpus that contain one of
the seed words. Each sentence is given to our parser, which segments
the sentence into simple noun phrases, verb phrases, and prepositional
phrases. For our purposes, we do not need any higher level parse
structures.

\item We collect small context windows surrounding each occurrence of
a seed word as a head noun in the corpus. Restricting the seed words
to be head nouns ensures that the seed word is the main concept of the
noun phrase. Also, this reduces the chance of finding different word
senses of the seed word (though multiple noun word senses may still be
a problem). We use a very narrow context window consisting of only two
words, the first noun to the word's right and the first noun to its
left. We collected only nouns under the assumption that most, if not
all, true category members would be nouns.\footnote{Of course, this
may depend on the target categories.} The context windows do
not cut across sentence boundaries. Note that our context window is
much narrower than those used by other
researchers~\cite{yarowsky92}. We experimented with larger window
sizes and found that the narrow windows more consistently included
words related to the target category.

\item
Given the context windows for a category, we compute a category score
for each word, which is essentially the conditional probability that
the word appears in a category context. The category score of a word W
for category C is defined as:
\begin{center}
{\it Score(W,C) = 
$\frac{freq.~of~W~in~C's~context~windows}{freq.~of~W~in~corpus}$}
\end{center}
Note that this is not exactly a conditional probability because a
single word occurrence can belong to more than one context window. For
example, consider the sentence: {\it I bought an AK-47 gun and
an M-16 rifle.} The word {\it M-16} would be in the context windows
for both {\it gun} and {\it rifle} even though there was just
one occurrence of it in the sentence. Consequently, the category
score for a word can be greater than 1. 

\item Next, we remove stopwords, numbers, and any words with a corpus
frequency $\leq$ 5. We used a stopword list containing about 30
general nouns, mostly pronouns (e.g., {\it I, he, she, they}) and
determiners (e.g., {\it this, that, those}). The stopwords and numbers
are not specific to any category and are common across many domains,
so we felt it was safe to remove them. The remaining nouns are sorted
by category score and ranked so that the nouns most strongly
associated with the category appear at the top.

\item The top five nouns that are not already seed words are added to
the seed word list dynamically. We then go back to Step 1 and repeat
the process. This bootstrapping mechanism dynamically grows the seed
word list so that each iteration produces a larger category
context. In our experiments, the top five nouns were added
automatically without any human intervention, but this sometimes
allows non-category words to dilute the growing seed word list. A few
inappropriate words are not likely to have much impact, but many
inappropriate words or a few highly frequent words can weaken
the feedback process. One could have a person verify that each word
belongs to the target category before adding it to the seed word list,
but this would require human interaction at each iteration of the
feedback cycle. We decided to see how well the technique could work
without this additional human interaction, but the potential benefits
of human feedback still need to be investigated.

\end{enumerate}

After several iterations, the seed word list typically contains many
relevant category words. But more importantly, the ranked list
contains many additional category words, especially near the top. The
number of iterations can make a big difference in the quality of the
ranked list. Since new seed words are generated dynamically without
manual review, the quality of the ranked list can deteriorate rapidly
when too many non-category words become seed words. In our
experiments, we found that about eight iterations usually worked well.

The output of the system is the ranked list of nouns after the final
iteration. The seed word list is thrown away. Note that the original
seed words were already known to be category members, and the new seed
words are already in the ranked list because that is how they were
selected.\footnote{It is possible that a word may be near the top of
the ranked list during one iteration (and subsequently become a seed
word) but become buried at the bottom of the ranked list during later
iterations. However, we have not observed this to be a problem so
far.}

Finally, a user must review the ranked list and identify the words
that are true category members. How one defines a ``true'' category
member is subjective and may depend on the specific application, so we
leave this exercise to a person. Typically, the words near the top of
the ranked list are highly associated with the category but the
density of category words decreases as one proceeds down the list.
The user may scan down the list until a sufficient number of category
words is found, or as long as time permits.  The words selected by
the user are added to a permanent semantic lexicon with the
appropriate category label.

Our goal is to allow a user to build a semantic lexicon for one or
more categories using only a small set of known category members as
seed words and a text corpus. The output is a ranked list of potential
category words that a user can review to create a semantic lexicon
quickly. The success of this approach depends on the quality of the
ranked list, especially the density of category members near the
top. In the next section, we describe experiments to evaluate our
system.

\section{Experimental Results}
\label{experiment-section}

We performed experiments with five categories to evaluate the
effectiveness and generality of our approach: {\it energy}, {\it
financial}, {\it military}, {\it vehicles}, and {\it weapons}. The
MUC-4 development corpus (1700 texts) was used as the text
corpus~\cite{muc4-proceedings}.  We chose these five categories
because they represented relatively different semantic classes, they
were prevalent in the MUC-4 corpus, and they seemed to be useful
categories.

For each category, we began with the seed word lists shown in
Figure~\ref{seed-words}. We ran the bootstrapping algorithm for eight
iterations, adding five new words to the seed word list after each
cycle. After the final iteration, we had ranked lists of potential
category words for each of the five categories. The top 45
words\footnote{Note that some of these words are not nouns, such as
{\it boarded} and {\it U.S.-made}. Our parser tags unknown words as
nouns, so sometimes unknown words are mistakenly selected for context
windows.}  from each ranked list are shown in
Figure~\ref{top-ranked-words}.

\begin{figure}[hbtp]
\fbox{
\begin{minipage}{2.9in}
{\bf Energy:} Limon-Covenas\footnote{Limon-Covenas refers to an oil
pipeline.} oligarchs spill staples poles Limon
Barrancabermeja Covenas 200,000 barrels oil Bucaramanga pipeline
prices electric pipelines towers Cano substation transmission rates
pylons pole infrastructure transfer gas fuel sale lines companies
power tower price gasoline industries insurance Arauca stretch inc
industry forum nationalization supply electricity controls
\rule{2.9in}{.01in}
{\bf Financial:} monetary fund nationalization \linebreak
attractive~circulation suit gold branches manager \linebreak bank
advice invested banks bomb\_explosion \linebreak investment invest
announcements content \linebreak managers insurance dollar savings
product \linebreak employee accounts goods currency reserves
\linebreak amounts money shops farmers maintenance \linebreak Itagui
economies companies foundation \linebreak moderation promotion
annually cooperatives \linebreak empire loans industry possession \\
\rule{2.9in}{.01in}
{\bf Military:} infantry 10th 3rd 1st brigade technician 2d 3d moran
6th 4th Gaspar 5th 9th Amilcar regiment sound 13th Pineda brigades
Anaya \linebreak division Leonel contra anniversary ranks \linebreak
Uzcategui brilliant Aristides escort dispatched \linebreak 8th Tablada employee
skirmish puppet \linebreak Rolando columns (FMLN) deserter 
troops \linebreak Nicolas Aureliano Montes Fuentes \\ \rule{2.9in}{.01in}
{\bf Vehicle:} C-47 license A-37 crewmen plate plates crash push tank
pickup Cessna aircraft cargo passenger boarded Boeing\_727 luxury
Avianca dynamite\_sticks hostile passengers accident sons airplane
light plane flight U.S.-made weaponry truck airplanes gunships fighter
carrier apartment schedule flights observer tanks planes
La\_Aurora\footnote{La\_Aurora refers to an airport.} fly helicopters
helicopter pole \\ \rule{2.9in}{.01in}
{\bf Weapon:} fragmentation sticks cartridge AK-47 M-16 carbines AR-15
movie clips knapsacks calibers TNT rifles cartridges theater 9-mm
40,000 quantities grenades machineguns dynamite kg ammunition
revolvers FAL rifle clothing boots materials submachineguns M-60
pistols pistol M-79 quantity assault powder fuse grenade caliber squad
mortars explosives gun 2,000
\end{minipage}
}

\caption{The top-ranked words for each category}
\label{top-ranked-words}
\end{figure}

While the ranked lists are far from perfect, one can see that there
are many category members near the top of each list. It is also
apparent that a few additional heuristics could be used to remove many
of the extraneous words. For example, our number processor failed to
remove numbers with commas (e.g., {\it 2,000}).  And the military
category contains several ordinal numbers (e.g., {\it 10th 3rd 1st})
that could be easily identified and removed . But the key question is
whether the ranked list contains many true category members. Since
this is a subjective question, we set up an experiment involving human
judges.

For each category, we selected the top 200 words from its ranked list
and presented them to a user. We presented the words in random order
so that the user had no idea how our system had ranked the words. This
was done to minimize contextual effects (e.g., seeing five category
members in a row might make someone more inclined to judge the next
word as relevant). Each category was judged by two people
independently.\footnote{The judges were members of our research group
but not the authors.}

The judges were asked to rate each word on a scale from 1 to 5
indicating how strongly it was associated with the category.  Since
category judgements can be highly subjective, we gave them guidelines
to help establish uniform criteria. The instructions that were given
to the judges are shown in Figure~\ref{instructions}.

\begin{figure}[hbtp]
\small
\fbox{
\begin{minipage}{2.9in}
CRITERIA: On a scale of 0 to 5, rate each word's strength of
association with the given category using the following
criteria. We'll use the category ANIMAL as an example. 
\begin{description}
\item[5:] CORE MEMBER OF THE CATEGORY: \\ If a word is clearly a member
of the category, then it deserves a 5. For example, dogs and sparrows
are members of the ANIMAL category. 

\item[4:] SUBPART OF MEMBER OF THE \\ CATEGORY: \\ If a word refers to a
part of something that is a member of the    category, then it
deserves a 4. For example, feathers and tails are    parts of ANIMALS. 

\item[3:] STRONGLY ASSOCIATED WITH THE \\ CATEGORY: \\ If a word refers to
something that is strongly associated with  members of the category,
but is not actually a member of the   category itself, then it
deserves a 3. For example, zoos and    nests are strongly associated
with ANIMALS.  
 
\item[2:] WEAKLY ASSOCIATED WITH THE \\ CATEGORY: \\ If a word refers to
something that can be associated with members    of the category, but
is also associated with many other types of   things, then it deserves
a 2. For example, bowls and parks are    weakly associated with ANIMALS. 

\item[1:] NO ASSOCIATION WITH THE CATEGORY: \\If a word has virtually no
association with the category, then it    deserves a 1. For example,
tables and moons have virtually no    association with ANIMALS. 

\item[0:] UNKNOWN WORD: \\ If you do not know what a word means, then it
should be labeled  with a 0. 
\end{description}
IMPORTANT! Many words have several distinct meanings. For example, the
word ``horse'' can refer to an animal, a piece of gymnastics equipment,
or it can mean to fool around (e.g., ``Don't horse around!'').  If a word
has ANY meaning associated with the given category, then only consider
that meaning when assigning numbers. For example, the word ``horse''
would be a 5 because one of its meanings refers to an ANIMAL.
\end{minipage}
}
\caption{Instructions to human judges}
\label{instructions}
\end{figure}

We asked the judges to rate the words on a scale from 1 to 5 because
different degrees of category membership might be acceptable for
different applications. Some applications might require strict
category membership, for example only words like {\it gun}, {\it
rifle}, and {\it bomb} should be labeled as weapons. But from a
practical perspective, subparts of category members might also be
acceptable. For example, if a {\it cartridge} or {\it trigger} is
mentioned in the context of an event, then one can infer that a gun
was used. And for some applications, any word that is strongly
associated with a category might be useful to include in the semantic
lexicon. For example, words like {\it ammunition} or {\it bullets} are
highly suggestive of a weapon.  In the UMass/MUC-4 information
extraction system~\cite{muc4-system}, the words {\it ammunition} and
{\it bullets} were defined as weapons, mainly for the purpose of
selectional restrictions.

The human judges estimated that it took them approximately 10-15
minutes, on average, to judge the 200 words for each category. Since
the instructions allowed the users to assign a zero to a word if they
did not know what it meant, we manually removed the zeros and assigned
ratings that we thought were appropriate. We considered ignoring the
zeros, but some of the categories would have been severely
impacted. For example, many of the legitimate weapons (e.g., M-16 and
AR-15) were not known to the judges. Fortunately, most of the unknown
words were proper nouns with relatively unambiguous semantics, so we
do not believe that this process compromised the integrity of the
experiment.

\begin{figure}[hbt]
\hsize=3in 
\epsffile{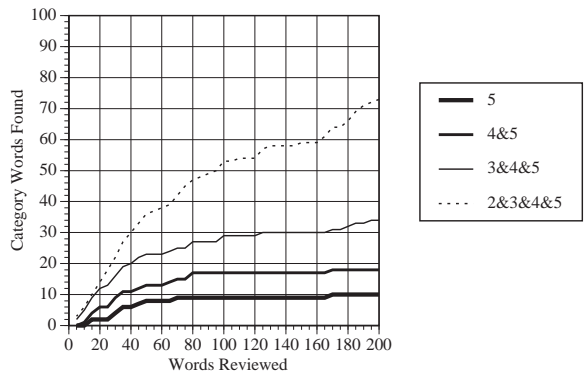}
\caption{Energy Results}
\label{energy-graph}
\end{figure}

\begin{figure}[hbt]
\centering 
\hsize=3in 
\epsffile{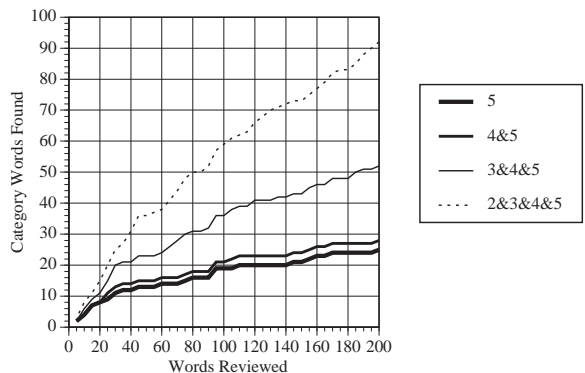}
\caption{Financial Results}
\label{financial-graph}
\end{figure}

\begin{figure}[hbt]
\centering 
\hsize=3in 
\epsffile{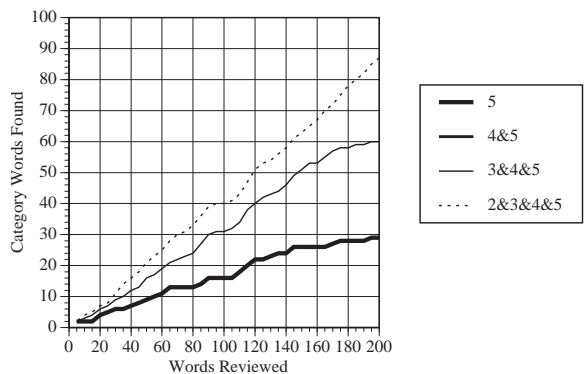}
\caption{Military Results}
\label{military-graph}
\end{figure}

\begin{figure}[hbt]
\centering 
\hsize=3in 
\epsffile{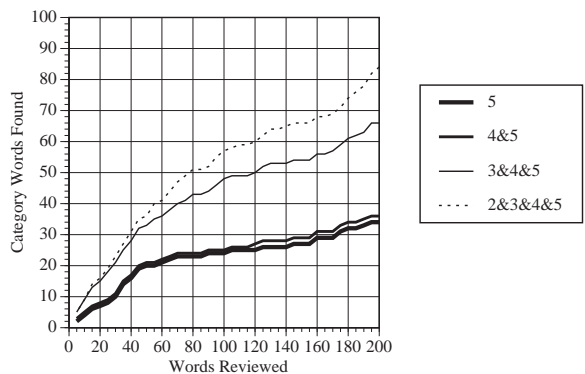}
\caption{Vehicle Results}
\label{vehicle-graph}
\end{figure}

\begin{figure}[hbt]
\centering 
\hsize=3in 
\epsffile{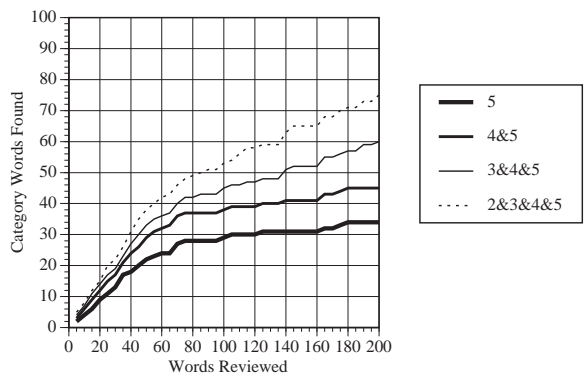}
\caption{Weapon Results}
\label{weapon-graph}
\end{figure}

Finally, we graphed the results from the human judges. We counted the
number of words judged as 5's by either judge, the number of words
judged as 5's or 4's by either judge, the number of words judged as
5's, 4's, or 3's by either judge, and the number of words judged as
either 5's, 4's, 3's, or 2's.  We plotted the results after each 20
words, stepping down the ranked list, to see whether the words near
the top of the list were more highly associated with the category than
words farther down. We also wanted to see whether the number of
category words leveled off or whether it continued to grow.  The
results from this experiment are shown in
Figures~\ref{energy-graph}-\ref{weapon-graph}.


With the exception of the Energy category, we were able to find 25-45
words that were judged as 4's or 5's for each category. This was our
strictest test because only true category members (or subparts of true
category members) earned this rating. Although this might not seem
like a lot of category words, 25-45 words is enough to produce a
reasonable core semantic lexicon. For example, the words judged as 5's
for each category are shown in Figure~\ref{5-words}. 

\begin{figure}[hbt]
\fbox{
\begin{minipage}{2.9in}
{\bf Energy:}  oil electric gas fuel power gasoline
electricity petroleum energy CEL \\
\rule{2.9in}{.01in}
{\bf Financial:} monetary fund gold bank invested banks investment
invest dollar currency money economies loans billion debts millions
IMF commerce wealth inflation million market funds dollars debt \\
\rule{2.9in}{.01in}
{\bf Military:} infantry brigade regiment brigades division ranks
deserter troops commander corporal GN Navy Bracamonte soldier units
patrols cavalry detachment officer patrol garrisons army paratroopers
Atonal garrison battalion unit militias lieutenant \\
\rule{2.9in}{.01in}
{\bf Vehicle:} C-47 A-37 tank pickup Cessna aircraft Boeing\_727
airplane plane truck airplanes gunships fighter carrier tanks planes
La\_Aurora helicopters helicopter automobile jeep car boats trucks
motorcycles ambulances train buses ships cars bus ship vehicle 
vehicles \\
\rule{2.9in}{.01in}
{\bf Weapon:}  AK-47 M-16 carbines AR-15 TNT rifles 9-mm grenades
machineguns dynamite revolvers rifle submachineguns M-60 pistols
pistol M-79 grenade mortars gun mortar submachinegun cannon RPG-7
firearms guns bomb machinegun weapons car\_bombs car\_bomb artillery
tanks arms
\end{minipage}
}
\caption{Words judged as 5's for each category}
\label{5-words}
\end{figure}

Figure~\ref{5-words} illustrates an important benefit of the
corpus-based approach. By sifting through a large text corpus, the
algorithm can find many relevant category words that a user would
probably not enter in a semantic lexicon on their own. For example,
suppose a user wanted to build a dictionary of Vehicle words. Most
people would probably define words such as {\it car}, {\it truck},
{\it plane}, and {\it automobile}. But it is doubtful that most people
would think of words like {\it gunships}, {\it fighter}, {\it carrier},
and {\it ambulances}.  The corpus-based algorithm is especially good
at identifying words that are common in the text corpus even though
they might not be commonly used in general. As another example,
specific types of weapons (e.g., {\it M-16}, {\it AR-15}, {\it M-60},
or {\it M-79}) might not even be known to most users, but they are
abundant in the MUC-4 corpus.

If we consider all the words rated as 3's, 4's, or 5's, then we were
able to find about 50-65 words for every category except Energy. Many
of these words would be useful in a semantic dictionary for the
category. For example, some of the words rated as 3's for the Vehicle
category include: {\it flight}, {\it flights}, {\it aviation}, {\it
pilot}, {\it airport}, and {\it highways}.

Most of the words rated as 2's are not specific to the target
category, but some of them might be useful for certain tasks. For
example, some words judged as 2's for the Energy category are: {\it
spill}, {\it pole}, {\it tower}, and {\it fields}. These words may
appear in many different contexts, but in texts about Energy topics
these words are likely to be relevant and probably should be defined
in the dictionary. Therefore we expect that a user would likely keep
some of these words in the semantic lexicon but would probably be
very selective.

Finally, the graphs show that most of the acquisition curves displayed
positive slopes even at the end of the 200 words. This implies that
more category words would likely have been found if the users had
reviewed more than 200 words. The one exception, again, was the Energy
category, which we will discuss in the next section. The size of the
ranked lists ranged from 442 for the financial category to 919 for the
military category, so it would be interesting to know how many
category members would have been found if we had given the entire
lists to our judges.

\section{Selecting Categories and Seed Words}
\label{categories-section}

When we first began this work, we were unsure about what types of
categories would be amenable to this approach. So we experimented with
a number of different categories. Fortunately, most of them worked
fairly well, but some of them did not. We do not claim to understand
exactly what types of categories will work well and which ones will
not, but our early experiences did shed some light on the strengths
and weaknesses of this approach.

In addition to the previous five categories, we also experimented with
categories for Location, Commercial, and Person. The Location category
performed very well using seed words such as {\it city}, {\it town},
and {\it province}. We didn't formally evaluate this category because
most of the category words were proper nouns and we did not expect
that our judges would know what they were. But it is worth noting that
this category achieved good results, presumably because location names
often cluster together in appositives, conjunctions, and nominal
compounds.

For the Commercial category, we chose seed words such as {\it store},
{\it shop}, and {\it market}. Only a few new commercial words were
identified, such as {\it hotel} and {\it restaurant}. In retrospect,
we realized that there were probably few words in the MUC-4 corpus
that referred to commercial establishments. (The MUC-4 corpus mainly
contains reports of terrorist and military events.) The relatively
poor performance of the Energy category was probably due to the same
problem.  If a category is not well-represented in the corpus then it
is doomed because inappropriate words become seed words in the early
iterations and quickly derail the feedback loop.

The Person category produced mixed results. Some good category words
were found, such as {\it rebel}, {\it advisers}, {\it criminal}, and
{\it citizen}. But many of the words referred to organizations (e.g.,
{\it FMLN}), groups (e.g., {\it forces}), and actions (e.g., {\it
attacks}). Some of these words seemed reasonable, but it was hard to
draw a line between specific references to people and concepts like
organizations and groups that may or may not consist entirely of
people. The large proportion of action words also diluted the
list. More experiments are needed to better understand whether this
category is inherently difficult or whether a more carefully chosen
set of seed words would improve performance.

More experiments are also needed to evaluate different seed word
lists. The algorithm is clearly sensitive to the initial seed words,
but the degree of sensitivity is unknown. For the five categories
reported in this paper, we arbitrarily chose a few words that were
central members of the category. Our initial seed words worked well
enough that we did not experiment with them very much. But we did
perform a few experiments varying the number of seed words. In
general, we found that additional seed words tend to improve
performance, but the results were not substantially different using
five seed words or using ten. Of course, there is also a law of
diminishing returns: using a seed word list containing 60 category
words is almost like creating a semantic lexicon for the category by
hand!

\section{Conclusions}
\label{conclusion-section}

Building semantic lexicons will always be a subjective process, and
the quality of a semantic lexicon is highly dependent on the task for
which it will be used.  But there is no question that semantic
knowledge is essential for many problems in natural language
processing. Most of the time semantic knowledge is defined manually
for the target application, but several techniques have been developed
for generating semantic knowledge automatically. Some systems learn
the meanings of unknown words using expectations derived from other
word definitions in the surrounding context
(e.g.,~\cite{granger77,carbonell79,jacobs88,hastings94}).  Other
approaches use example or case-based methods to match unknown word
contexts against previously seen word contexts
(e.g.,~\cite{berwick89,cardie93}). Our task orientation is a bit
different because we are trying to construct a semantic lexicon for a
target category, instead of classifying unknown or polysemous words in
context.

To our knowledge, our system is the first one aimed at building
semantic lexicons from raw text without using any additional semantic
knowledge. The only lexical knowledge used by our parser is a
part-of-speech dictionary for syntactic processing. Although we used a
hand-crafted part-of-speech dictionary for these experiments,
statistical and corpus-based taggers are readily available
(e.g.,~\cite{brill94,church89,weischedel93}).

Our corpus-based approach is designed to support fast semantic lexicon
construction.  A user only needs to supply a representative text
corpus and a small set of seed words for each target category. Our
experiments suggest that a core semantic lexicon can be built for each
category with only 10-15 minutes of human interaction. While more work
needs to be done to refine this procedure and characterize the types
of categories it can handle, we believe that this is a promising
approach for corpus-based semantic knowledge acquisition.

\section{Acknowledgments} 
This research was funded by NSF grant IRI-9509820 and the University
of Utah Research Committee.  We would like to thank David Bean, Jeff
Lorenzen, and Kiri Wagstaff for their help in judging our category
lists.

\end{document}